\documentclass[accepted]{uai2023} 

\usepackage[american]{babel}

\usepackage{natbib} 
\usepackage{mathtools} 
\usepackage{booktabs} 
\usepackage{tikz} 

\usepackage{microtype}
\usepackage{graphicx}
\usepackage{subfigure}
\usepackage{booktabs} 

\usepackage{hyperref}

\usepackage{amsmath}
\usepackage{amssymb}
\usepackage{mathtools}
\usepackage{amsthm}

\usepackage[capitalize,noabbrev]{cleveref}

\theoremstyle{plain}

\theoremstyle{definition}

\theoremstyle{remark}

\usepackage{algorithm}
\usepackage{algorithmic}
\usepackage{textcase}
\usepackage{multirow}
\usepackage{bm}
\usepackage{xcolor}
\usepackage{tabularx}

\newcommand{\replay}{\mathcal{D}}
\newcommand{\jointaction}{\mathbf{a}}
\newcommand{\jointhist}{\boldsymbol{\tau}}
\newcommand{\statespace}{\mathcal{S}}
\newcommand{\actionspace}{\mathcal{A}}
\newcommand{\indparam}{\boldsymbol{\phi}}

\newcommand{\EE}{\mathbb{E}}
\newcommand{\qi}{Q_{i}}
\newcommand{\qjt}{Q_{joint}}
\newcommand{\supidp}[1]{#1^{idp}}
\newcommand{\supdep}[1]{#1^{dep}}
\newcommand{\subact}{\text{act}}
\newcommand{\subboot}{\text{boot}}
\newcommand{\subrew}{\text{rew}}
\newcommand{\mixer}{\text{Mixer}}
\newcommand{\rom}[1]{\romannumeral #1}

\usepackage{easyReview}

\title{Conditionally Optimistic Exploration for Cooperative Deep Multi-Agent Reinforcement Learning}

\author[1,2]{Xutong Zhao}
\author[3]{Yangchen Pan}
\author[4]{Chenjun Xiao}
\author[1,2,6]{Sarath Chandar}
\author[1,5]{Janarthanan Rajendran}
\affil[1]{
    Mila - Quebec AI Institute
}
\affil[2]{
    \'Ecole Polytechnique de Montr\'eal
}
\affil[3]{
    University of Oxford
  }
\affil[4]{
    University of Alberta
}
\affil[5]{
    Universit\'e de Montr\'eal
}
  
\begin{document}
\maketitle

\begin{abstract}\label{sec:abs}
Efficient exploration is critical in cooperative deep Multi-Agent Reinforcement Learning (MARL).
In this work, we propose an exploration method that effectively encourages cooperative exploration based on the idea of sequential action-computation scheme.
The high-level intuition is that to perform optimism-based exploration, agents would explore cooperative strategies if each agent's optimism estimate captures a structured dependency relationship with other agents.
Assuming agents compute actions following a sequential order at \textit{each environment timestep}, we provide a perspective to view MARL as tree search iterations by considering agents as nodes at different depths of the search tree.
Inspired by the theoretically justified tree search algorithm UCT (Upper Confidence bounds applied to Trees), we develop a method called Conditionally Optimistic Exploration (COE).
COE augments each agent's state-action value estimate with an action-conditioned optimistic bonus derived from the visitation count of the global state and joint actions of preceding agents.
COE is performed during training and disabled at deployment, making it compatible with any value decomposition method for centralized training with decentralized execution.
Experiments across various cooperative MARL benchmarks show that COE outperforms current state-of-the-art exploration methods on hard-exploration tasks.
\end{abstract}

\section{Introduction}\label{sec:intro}
In recent years multi-agent reinforcement learning (MARL) has drawn much attention and has shown high potential to be applied to various real-world scenarios, such as transportation \citep{seow2009collaborative}, robotics \citep{huttenrauch2017guided,perrusquia2021multi}, and autonomous driving \citep{cao2012overview,shalev2016safe}.
Cooperative MARL is concerned with a special learning setting with common rewards shared across all agents, where agents must coordinate their strategies to achieve a common goal.
There are several major challenges posed by this setting, such as credit assignment, scalability, non-stationarity, and partial observability.
To address these challenges, \citet{bernstein2002complexity} propose the Centralized Training with Decentralized Execution (CTDE) learning paradigm.
In this paradigm, information is shared across agents during training, guiding the learning of individual agent's policies and promoting cooperation during training, while agents still being able to run independently during decentralized execution.

One important line of research in CTDE is value decomposition,
which learns a centralized action-value function that can be factorized into the individual utility function (often referred to as individual Q-function) of each agent.
The centralized value function performs implicit credit assignment to determine each agent's contribution towards common returns, and learns implicit inter-dependencies to encourage cooperation.
To ensure the centralized policy is aligned with individual policies, \citet{son2019qtran} propose the Individual-Global-Max (IGM) principle that guarantees consistency between global and local greedy actions.
A common approach to value decomposition is to learn a mixing network that computes the centralized action value from the utilities of all agents.
Depending on the specific way to satisfy IGM, different methods have been introduced, including VDN \citep{sunehag2017value}, QMIX \citep{rashid2018qmix}, QTRAN \citep{son2019qtran}, and QPLEX \citep{wang2020qplex}.

Cooperative exploration adds another level of difficulty to the single-agent exploration challenge.
In cooperative MARL, agents need to coordinate to explore the large joint state-action space as high-performing joint strategies may require a high degree of collaboration among them.
In addition, there may exist different types of cooperative strategies associated with a task.
Sound cooperative exploration methods should be able to identify the optimal strategy from potentially many sub-optimalities.
For instance, if the task for a group of robots is to deliver packages in a warehouse, the optimal strategy is that at each time several agents together lift a heavy package that one single agent cannot lift, meanwhile, each other spare agent carries one small package.
In this task, delivering items either only collectively or only separately is sub-optimal, even though either way agents achieve a cooperative strategy.
Therefore cooperative exploration is challenging in cooperative MARL.
Although directed exploration strategies have been widely studied in multi-armed bandit and single-agent RL settings, they fail to account for cooperation among agents.
Moreover, it is not straightforward to adopt single-agent methods to cooperative MARL, due to the exponentially large state-action space and multi-agent credit assignment.
The popular $\varepsilon$-greedy strategy has been shown to be ineffective in complex MARL coordination tasks \citep{wang2020qplex}.

Some recent works encourage cooperative exploration in MARL settings by maximizing the correlation among agents' behaviour, which trains each agent's policy to account for influences from other agents, hence agents achieve effective collaborative exploration behaviour.
Correlation maximization is often realized by maximizing the mutual information (MI) between some quantities that can determine or reflect agents' behaviours,
such as the trajectory history of each agent.
Utilizing this idea, some works have been proposed and empirically outperformed the value decomposition baselines across various benchmark tasks \citep{jaques2019social,mahajan2019maven,wang2019influence,kim2020maximum,li2021celebrating}.
However, two major issues remains when MI-based methods are used.
First, optimizing the MI quantity for every pair of agents is not scalable because the required computation to optimize all MI losses grows quadratically as the number of agents.
Second, agents could learn different types of cooperative strategies, and one particular type may not lead to high performance.
As pointed out by \citet{li2022pmic}, simply maximizing the MI may not lead to high returns because agents may learn sub-optimal joint strategy, regardless of how strong the correlation they achieve.

In this work, we seek to explicitly leverage inter-agent dependencies to drive cooperative exploration.
Our insight is simple: as a complement to implicit dependencies learned by centralized training, if each agent's optimism estimate explicitly encodes a structured dependency relationship with other agents, by performing optimism-based exploration, agents would be guided to effectively explore cooperative strategies.
It is worth noting that centralized training (CT) only requires the joint policy to output the joint actions of all agents at the same time, without posing restrictions on how the controlling algorithm performs internal calculations or what information is allowed to be shared across agents.
During CT, assuming at \textit{each environment timestep} agents compute actions according to a sequential pre-determined order before executing them simultaneously, we can view the action computation sequence as a path from the root to a leaf of a tree.
At each node of the tree, we consider the preceding agent's action as the parent node of the current agent.
We revisit the idea of UCT exploration \citep{kocsis2006bandit} proposed in the perfect-information game setting, where the game state is accessible at all nodes, and take inspiration from it to encourage cooperative exploration in MARL.
We develop a method called Conditionally Optimistic Exploration (COE).
Essentially, COE performs optimism-based exploration by computing the upper confidence bounds of each action for the current agent, conditioned on the visitation count of its parent node (i.e., preceding agents' actions).
To obtain decentralized agents in the decentralized execution (or deployment) phase, COE is not applied, i.e., we disable exploration by removing the optimistic bonus terms.

In the subsequent sections,
we first review the background on MARL and the UCT algorithm. 
We then describe how conditional optimism can be applied to the MARL setting to encourage cooperative exploration.
We build COE on commonly used value decomposition methods, and utilize the hash-based counting technique \citep{tang2017exploration} to enable counting the visitations in continuous state-action domains.
Our empirical results on various benchmark domains show that our method is more effective than well-known baselines in challenging exploration tasks, and matches baseline performance in general MARL tasks.
Our source code is available at \url{https://github.com/chandar-lab/COE}.

\section{Background}\label{sec:background}

\subsection{Dec-POMDP}
We model the cooperative multi-agent task as a Dec-POMDP (Decentralized Partially Observable Markov Decision Process) \citep{oliehoek2016concise}, which is formally defined as a tuple $G = \langle \statespace, \actionspace, P, R, \Omega, O, n, \gamma \rangle$, where $\statespace$ is the global state space, $\actionspace$ is the action space, $\Omega$ is the observation space, $n$ is the number of agents in the environment, and $\gamma \in [0,1]$ is the discount factor. 
At each timestep $t$, on state $s \in \statespace$ each agent $i \in \mathcal{N} \equiv \{1,\dots,n\}$ takes an action $a_i \in \actionspace$. The joint action  $\mathbf{a} = [a_i]^n_{i=1} \in \boldsymbol{\actionspace} \equiv \actionspace^n$ leads to the next state $s'$ sampled from the transition probability $P(s'|s, \mathbf{a}): \statespace \times \boldsymbol{\actionspace} \times \statespace \rightarrow [0,1]$, and obtains a global reward $r$ according to the reward function $R(s, \mathbf{a}): \statespace \times \boldsymbol{\actionspace} \rightarrow \mathbb{R}$ shared across all agents.
Each agent $i$ has a local policy $\pi_i(a_i|s): \statespace \times \actionspace \rightarrow [0, 1]$.
Based on the joint policy $\boldsymbol{\pi} \equiv [\pi_i]^n_{i=1}$, the joint action-value function is defined as $Q_{\boldsymbol{\pi}}(s, \mathbf{a}) = \EE_{\boldsymbol{\pi}} \left[ \sum^\infty_{k=0} \gamma^{k} r^{(t+k)}| s^{(t)} = s, \mathbf{a}^{(t)} = \mathbf{a} \right]$. The objective is to find a joint policy that maximizes the action-value function.

We consider the partially observable setting, where each agent $i$ does not observe the global state $s$, instead only has access to a local observation $o_i \in \Omega$ drawn from the observation function $O(s, i): \statespace \times \mathcal{N} \rightarrow \Omega$.
Hence each agent $i$ maintains its action-observation history $\tau_i \in T \equiv (\Omega \times \actionspace)^* \times \Omega$, on which it can condition its policy $\pi_i(a_i|\tau_i): T \times \actionspace \rightarrow [0, 1]$. 
With agent $i$ observing the next observation $o_i'$, the updated next history is represented by $\tau_i' = \tau_i \cup \{a_i, o_i' \}$.
We denote the joint history by $\jointhist \equiv [\tau]^n_{i=1} \in \mathbf{T} \equiv T^n$, and similarly joint next history by $\jointhist' \equiv [\tau']^n_{i=1}$.

\subsection{UCT}
UCT (Upper Confidence bounds applied to Trees) \citep{kocsis2006bandit} is a tree search algorithm commonly used in Monte-Carlo Tree Search for perfect-information games.
In UCT, node selection is treated as a multi-armed bandit problem, where at each node its children nodes correspond to the arms,
and the Upper Confidence Bound (UCB) bandit algorithm \citep{auer2002finite} is used to select the child node with the highest upper confidence.
In particular, consider a sequence of node selections from the root to a leaf of a search tree as a trajectory at one timestep, at each depth the child node $i$ with the highest upper confidence bound is selected:
\begin{equation}\label{eq:uct-bandit-act}
    B_i = X_i + c \sqrt{\frac{2 \log(p)}{n_i}},
\end{equation}
where $X_i$ is the empirical mean of the rewards that have been obtained by trajectories going through node $i$,
$c$ is a constant controlling the scale of exploration,
$n_i$ and $p$ are the number of times node $i$ and its parent node have been visited, respectively.
Intuitively, conditioned on previously taken actions in the trajectory, at the current node actions that have been taken fewer times will have a higher exploration bonus, hence UCT tends to take action combinations that are under-explored or promising actions with higher reward estimates.
When the trajectory is completed, a reward is received at the leaf. 
The visitation count and reward estimate of each selected node are updated accordingly.
The original paper provides a regret analysis of the UCT algorithm, proving that its expected regret is upper bounded by $O(\log t)$, where $t$ is the number of trajectories/timesteps.

\section{Related Work}\label{sec:related}
\paragraph{Single-agent exploration.}
Exploration strategies have been extensively studied in single-agent deep RL settings.
\citet{amin2021survey} provide a thorough literature survey of advanced exploration methods.
In recent years, the category of bonus-based methods has been commonly applied to solve hard exploration tasks.
Based on the Optimism in the Face of Uncertainty (OFU) principle, the high-level idea is to capture some notion of uncertainty (often referred to as novelty or curiosity), and augment the extrinsic reward the environment emits with an intrinsic reward that quantifies the uncertainty.
For instance, the count-based method \citep{bellemare2016unifying,ostrovski2017count,tang2017exploration} measures novelty through the number of times the agent observes a state-action tuple.
The learning progress methods (such as \citet{houthooft2016vime,pathak2017curiosity,burda2018exploration,pathak2019self}) capture the curiosity through the learning progress of the agent's knowledge of the environment.
Despite their state-of-the-art performance on hard exploration tasks in the Atari benchmark \citep{taiga2019benchmarking},
naively applying intrinsic reward methods to MARL may not work well due to the multi-agent credit assignment challenge.
As the same global reward signal is shared across all agents, training local policies still relies on the centralized value/critic function to perform implicit credit assignment; hence augmenting an intrinsic reward may still be inefficient to learn structured exploration and cooperation.
Other successful exploration approaches like BootstrappedDQN \citep{osband2016deep} or Go-Explore \citep{ecoffet2021first} are unscalable in MARL due to the exponentially large state-action space.

\paragraph{Multi-agent exploration.}
A recent branch of research proposes to drive multi-agent exploration by promoting collaboration among agents through the maximization of the correlation or influence among agents.
The correlation is commonly realized by the mutual information (MI) of quantities that define or reflect agents' behaviour,
such as the trajectory history of each agent.
For instance, MAVEN \citep{mahajan2019maven} learns a hierarchical policy to produce a latent variable that encodes the information about the joint policy, and maximizes the MI between this latent variable and the joint trajectories to encourage the correlation of agents' behaviour. 
Some other methods try to promote collaboration by maximizing pairwise MI between every two agents in the form of intrinsic rewards.
For instance, EITI \citep{wang2019influence} maximizes the MI between one agent's state transition and the other's state-action.
VM3-AC \citep{kim2020maximum} maximizes the MI between two agents' policy distributions.
Pairwise MI is hard to scale to scenarios with a large number of agents, because the computation grows quadratically with the number of agents.
Moreover, as mentioned in the previous paragraph about single-agent exploration, the same multi-agent credit assignment challenge persists in intrinsic reward MARL methods in general given a centralized value/critic function is used.
\citet{li2022pmic} claims one other important downside of MI-based methods is the fact that a strong correlation does not necessarily correspond to high-return collaboration, especially when there exist multiple sub-optimal highly-cooperative strategies associated with the given task.
Aside from MI-based methods, there are other approaches based on different intuitions.
VACL \citep{chen2021variational} leverages variational inference and automatic curriculum learning to solve sparse-reward cooperative MARL challenges.
EMC \citep{zheng2021episodic} utilizes a value decomposition method to implicitly capture influence among agents and uses the prediction errors of individual Q-value functions as intrinsic rewards.
EMC achieves state-of-the-art performance on multiple challenging tasks in the StarCraft Multi-Agent Challenge \citep{samvelyan2019starcraft} benchmark.
Our method also tries to capture agent-wise dependency to guide exploration.
Different from MI maximization or EMC, our method captures structured inter-dependency through each agent's conditional optimism estimate and performs optimism-based exploration.

\paragraph{Action Conditioned Learning.}
As the learning objective in CTDE is to obtain decentralized agents for execution, previous works commonly assume agents both compute and take actions simultaneously, even during the centralized training phase.
A few recent works explicitly consider inter-dependency and cooperation learned through sequential action computation, where at each timestep each agent's policy is conditioned on preceding agents' joint action.
MACPF \citep{wang2022more} learns a dependent joint policy and its independent counterpart by maximum-entropy RL \citep{ziebart2010modeling}.
ACE \citep{li2022ace} is a Q-learning method that models the multi-agent MDP into a single-agent MDP by making the bootstrap target dependent on subsequent agents' actions.
Leveraging the multi-agent advantage decomposition theorem \citep{kuba2021trust}, Multi-Agent Transformer (MAT) \citep{wen2022multi} casts MARL into a sequence modeling problem and uses a transformer architecture to map agents' observation sequences to agents' optimal action sequences.
These methods consider action conditioning to increase the expressiveness of the joint policy, hence improving its performance.
Our method leverages action conditioning from a different perspective: predecessors' actions reflect dependency among agents, therefore can be used to adjust the optimism level to achieve efficient cooperative exploration.

\section{Conditionally Optimistic Exploration}\label{sec:method}

In this section, we introduce our method Conditionally Optimistic Exploration (COE) that effectively drives exploration in cooperative deep MARL.
We describe how we can view cooperative MARL as a sequence of tree search iterations.
We then discuss how we take inspiration from the tree search method UCT, as well as the challenges to directly applying its idea to MARL.
We finally present approaches to address these challenges and the details of our proposed COE method.

\begin{figure}[!tb] 
  \centering
  \includegraphics[width=1.0\linewidth]{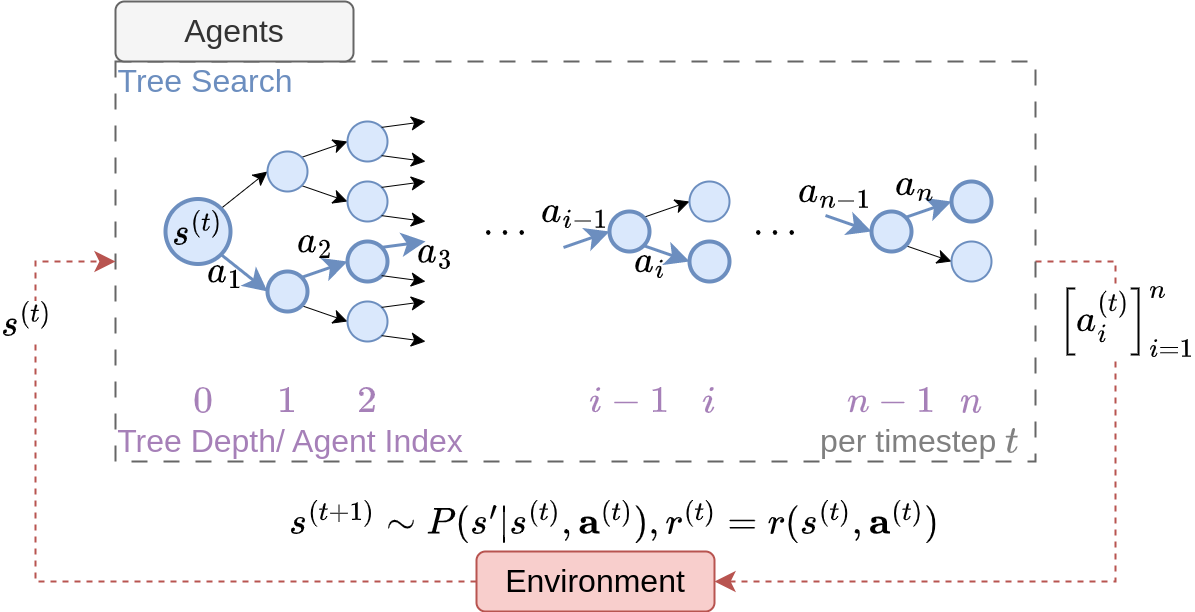}
  \caption{Modelling of MARL as Tree Search Procedure.}\label{fig:marl-as-tree}
\end{figure}

\subsection{MARL Action Computation as a Tree}
We first formulate action computation of MARL as a tree search procedure at each environment timestep.
We consider the following sequential decision scheme: at each timestep $t$, all $n$ agents \textit{compute their actions sequentially} following some arbitrary but fixed order,
and then \textit{execute these actions simultaneously} in the environment.
Without loss of generality, we use the identities of agents as the order, i.e., agent $i \in \mathcal{N} \equiv \{1,\dots,n\}$ is the $i$-th agent to select its action.
\cref{fig:marl-as-tree} depicts the formulation of MARL action computation as a tree search at each timestep $t$. For simplicity of illustration, the figure shows a case where each agent has to choose between two actions.
To construct the tree structure, we first consider the state $s^{(t)}$ as the root node.
Each agent $i$ is the $i$-th agent to choose its action,
then every node at depth $i>0$ represents an intermediate stage to which an action sequence of agents $\{1,\dots,i\}$ sequentially transitions.
Each agent $i$ has $|\actionspace| = k$ actions, corresponding to the $k$ children nodes of the parent node at depth $i-1$.
When agent $i$ chooses its action, the computation is conditioned on the joint action of agents $\{1,\dots,i-1\}$.
When agent $n$ computes its action, the action sequence reaches a leaf node as shown with bold arrows in the tree diagram, where all internal action computations are completed.
The agent-environment interaction is standard in Dec-POMDP: all $n$ agents execute/take their action simultaneously in the environment; upon receiving the joint action, the environment emits a reward to all agents, and transitions to the next state $s^{(t+1)}$, i.e., the root node of the next tree.

Having framed MARL action computation into a tree structure, to guide MARL exploration we could take inspiration from bandit-based tree search methods such as UCT.
By applying the idea of UCT to each tree, we model the cooperative MARL exploration as a sequence of conditional optimistic exploration procedures as follows.
For action selection at each state $s$, conditioned on its predecessors' joint action, denoted by $a_{<i}$, each agent $i$ estimates an action-value function $\supdep{\qi}(s, a_i | a_{<i})$, where the superscript $dep$ represents $\qi$'s dependence on $a_{<i}$.
Augmenting the Q-value by an optimistic bonus conditioned on preceding agents' actions, we obtain an optimistic action value estimate $\supdep{B_i}(s, a_i | a_{<i})$ to determine action selection.
It is worth noting that the Q-value estimate is generalizable across states and preceding agents' joint actions using neural network function approximators, which avoids maintaining an empirical reward estimate at every node in every tree.
At depth $i$ of each tree, agent $i$ uses the same Q-value function approximator to select action, no matter which subtree the corresponding node is in.

\begin{figure*}[!tb] 
  \centering
  \includegraphics[width=0.9\linewidth]{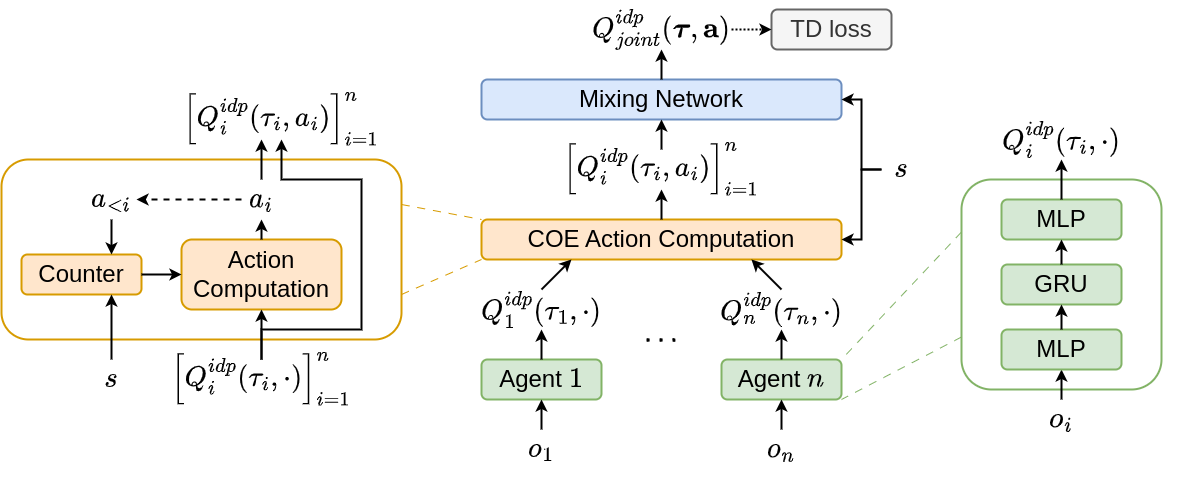}
  \caption{Learning Framework for COE. COE Performs Conditionally Optimistic Action Computation on Standard Value Decomposition Framework.}\label{fig:coe-framework}
\end{figure*}

It should be noted that directly applying tree-based algorithms is incompatible with the cooperative MARL setting due to its distinctions with the conventional tree search problem setting.
Tree search methods, especially those that rely on visitation count like UCT, typically assume the global state information is accessible at all nodes, whereas the Dec-POMDP setting assumes partial observability of each agent.
Accessing full state information enables agents to estimate action values based on the same global state and predecessors' actions, while in Dec-POMDP agents cannot have such estimates.
CTDE also requires agents to act independently at execution time, without conditioning policies on other agents' actions.
To tackle these challenges, we (1) develop an approximate implementation by building conditional optimism on a value decomposition method, and (2) disable exploration after training to obtain decentralized agents.
We present details of our approach in the next subsection,
and empirically evaluate it in \cref{sec:experiment}.

\subsection{COE Algorithm}
We first briefly describe the value decomposition learning paradigm \citep{sunehag2017value,rashid2018qmix}.
We then present how we utilize conditional counts on value decomposition to drive optimistic exploration.

Each agent $i$ has an independent Q-network $\supidp{\qi} (\tau_i, a_i; \phi_i)$ parameterized by $\phi_i$.
It is important to note that the superscript $idp$ indicates that the $\qi$ is \textit{independent} of other agents' actions, as opposed to a $\supdep{\qi}$ that is \textit{dependent} on predecessors' actions if action computation follows a sequential order.
The same naming rule also applies to joint Q-values.
A mixing network $\mixer(\cdot ; \theta)$ parameterized by $\theta$ is used to compute the joint Q-values from all individual Q-values:
\begin{equation}\label{eq:q-joint-ind}
    \supidp{\qjt}(\jointhist, \jointaction) = \mixer \left( \left[ \supidp{\qi} \left(\tau_i, a_i \right) \right]^n_{i=1}, s; \theta \right).
\end{equation}
Individual agent's action-value networks $\supidp{\qi}$ and the mixing network $\mixer$ are trained by minimizing the mean-squared temporal-difference error:
\begin{equation} 
    \supidp{\mathcal{L}} \left( \left[\phi \right]^n_{i=1}, \theta \right) = \EE_\replay \left[ \left(\supidp{\qjt} \left(\jointhist, \jointaction \right) - \supidp{y} \right)^2 \right] \label{eq:td-err-ind}
\end{equation}
where $\supidp{y} = \left(r + \gamma \max_{\jointaction'} \left(\supidp{\qjt} \left(\jointhist', \jointaction' \right) \right) \right)$ is the update target, and $\replay$ is the replay buffer containing trajectory data collected by all agents.
It is worth noting that by IGM principle, the greedy actions selected by $\supidp{\qi}$'s are the same actions $\supidp{\qjt}$ would have taken. 
As centralized training backpropagates the global reward signal to learn the individual utilities $\supidp{\qi}$'s, value factorization implements an implicit multi-agent credit assignment that enables each agent to grasp the inter-dependency among all utilities.

Building on top of the value decomposition skeleton, we incorporate count-based optimism in both action computation and learning during the centralized training (CT) phase.
For action computation, each agent $i$ selects greedy actions with respect to its conditional optimistic action-value
\begin{equation}\label{eq:ucb-act}
a_i = \arg\max_{a_i'}  \left\{  \supidp{\qi}  \left(\tau_i, a_i' \right) + c_\subact \sqrt{\frac{2 \log \left(N \left(s, a_{<i} \right)\right) }{N \left(s, a_{<i}, a_i' \right) }}  \right\},
\end{equation}
where $c_\subact \in \mathbb{R}_+$ is a hyper-parameter controlling the scale of optimism, $N(\cdot)$ denotes the visitation count.
Note that counting is performed in the global state space, thanks to centralized training.
The learning framework of COE is illustrated in \cref{fig:coe-framework}.

Moreover, we augment the global reward and the bootstrapped target each with a bonus term, such that the update target becomes
\begin{multline}\label{eq:ucb-target-joint}
\supidp{y} = \left( r \left(s, \jointaction \right) + \frac{c_\subrew}{\sqrt{N \left(s, \jointaction \right)}} \right) + \gamma \max_{\jointaction'}  \Biggr[ \\
    \mixer  \left(  \left[  \supidp{\qi}  \left(\tau_i', a_i' \right) + \frac{c_\subboot}{\sqrt{N \left(s', a_{<i}', a_i' \right)}}  \right]^n_{i=1}  \right) \Biggr],
\end{multline}
where $c_\subrew, c_\subboot \in \mathbb{R}_+$ are hyper-parameters controlling the scale of the optimistic bias in reward and bootstrapped target, respectively.
These two bonus terms are added for two major reasons.
First, we intend to maintain long-term optimism in the Q-functions.
The acting-time optimism decreases as the corresponding count is incremented, but unlike bandit or tabular MDP methods, COE's Q-value estimate is updated at a relatively slower rate due to the nature of gradient updates of neural networks.
To encourage COE to explore persistently, the augmentation to the bootstrap target allows the Q-value itself to encode optimism through TD loss update.
Second, since the bootstrap target is defined based on the Q-value estimates of the next state-actions, this optimistic bootstrap target also captures uncertainty from subsequent agents and future timesteps.
The idea of learning optimistic Q-values originates from theoretical works such as \citet{jin2018q,jin2020provably,yang2020function}, and has been extended to deep RL recently (e.g., \citet{rashid2020optimistic}).

With the count-based optimism introduced, the complete learning algorithm is presented in \cref{alg:uct-qlearning}.
During decentralized execution, the optimistic bonuses, although may have decayed to negligible magnitude, are removed, and agents take independent actions according to $\supidp{\qi}$'s only.

To apply COE to deep MARL tasks, we need to approximate counts in high-dimensional or continuous state space.
In our experiments, we use the SimHash method \citep{tang2017exploration} that projects states to a lower-dimensional feature space before counting.
We record the visitation count for the tuple of the state $s$ and all agents' joint action $\jointaction$, denoted by $N(s, \jointaction)$.
For each agent $i$, the count up to its action $a_i$ satisfies $N(s, a_{<i}, a_i) = \sum_{a_{i+1}} N(s, a_{<i}, a_i, a_{i+1}) = \sum_{a_{>i}} N(s, a_{<i}, a_i, a_{>i})$,
where $a_{<i}$ and $a_{>i}$ denote the joint actions computed by preceding and subsequent agents of $i$, respectively.
This relationship shows that we can obtain any count up to $a_i$ by summing up the counts of joint actions that share the same $a_{<i}$ at state $s$.
Details about SimHash counting are presented in 
\cref{apx:pseudo-count}. 

\begin{algorithm}[tb]   
   \caption{Conditionally Optimistic Exploration}
   \label{alg:uct-qlearning}
\begin{algorithmic}
    \STATE Initialize parameters $\indparam, \theta$
    \STATE Visitation count $N(s,\jointaction) \leftarrow 0, \forall (s, \jointaction) \in \statespace \times \boldsymbol{\actionspace}$
    \STATE Replay buffer $\replay \leftarrow \{\}$
    \FOR{each episode $m=1, \dots, M$}
        \FOR{each environment timestep $t=1, \dots, T$}
            \FOR{agent $i=1, \dots, n $}
                \STATE Compute action $a^{(t)}_i$ according to \cref{eq:ucb-act}
            \ENDFOR
            \STATE $s^{(t+1)} \sim P(s'|s^{(t)},\jointaction^{(t)}), r^{(t)} = r(s^{(t)},\jointaction^{(t)})$
            \STATE $N(s^{(t)}, \jointaction^{(t)}) \leftarrow N(s^{(t)}, \jointaction^{(t)}) + 1$
            \STATE $\replay \leftarrow \replay \cup \left\{ \left(s^{(t)}, \jointaction^{(t)}, r^{(t)}, s^{(t+1)} \right) \right\}$
            \STATE Perform a gradient update on \cref{eq:td-err-ind}
        \ENDFOR
    \ENDFOR
\end{algorithmic}
\end{algorithm}

\section{Experiments}\label{sec:experiment}
In this section, we evaluate COE on cooperative MARL tasks across three commonly used benchmarks: Multi-agent Particle Environments (MPE) \citep{lowe2017multi,mordatch2018emergence}, Level-Based Foraging (LBF) \citep{albrecht2015game,christianos2020shared,papoudakis2020benchmarking}, and StarCraft Multi-Agent Challenge (SMAC) \citep{samvelyan2019starcraft}.
These tasks can be categorized to two sets based on challenges they exhibit:
(1) sparse-reward tasks that specifically pose the cooperative exploration challenge,
and (2) tasks that generally assess MARL methods' ability for effective coordination.
Empirical results show that COE achieves higher sample efficiency and performance than other state-of-the-art approaches in sparse-reward tasks, and matches their performance in general cooperative tasks.
We also present ablation studies to demonstrate the effectiveness of conditional optimism and COE's compatibility with common MARL methods.
As a sanity check, we examine conditional optimism in a didactic repeated multi-player game problem.

\begin{table*}[!htb]
    \scriptsize
    \centering
    \caption{Average Returns and 95\% Confidence Interval for All Four Algorithms, and Average Win-rates for SMAC Tasks.}\label{tab:avg-return-baseline}
    \begin{tabular}{*{3}{l}*{4}{c}}
      \toprule
      && \multicolumn{1}{l}{\textbf{Tasks \textbackslash Algs.}} & \multicolumn{1}{c}{COE} & \multicolumn{1}{c}{EMC} & \multicolumn{1}{c}{MAVEN} & \multicolumn{1}{c}{QMIX} \\
      \midrule
      \multirow{3}{*}{\rotatebox[origin=c]{90}{MPE}}
      && Adversary & $17.77 \pm 0.71$ & $16.73 \pm 0.83$ & $\bm{19.57 \pm 0.51}$ & $18.20 \pm 0.56$ \\
      && \textbf{Sparse Tag} & $\bm{0.65 \pm 0.09}$ & $0.43 \pm 0.06$ & $0.01 \pm 0.00$ & $0.40 \pm 0.05$ \\
      && \textbf{Sparse Spread} & $\bm{0.79 \pm 0.09}$ & $0.41 \pm 0.18$ & $0.10 \pm 0.14$ & $0.29 \pm 0.05$ \\
      \midrule
      \multirow{4}{*}{\rotatebox[origin=c]{90}{LBF}}
      && \textbf{10x10-3p-3f} & $\bm{0.71 \pm 0.05}$ & $0.68 \pm 0.03${*} & $0.16 \pm 0.06$ & $0.49 \pm 0.01$ \\
      && \textbf{15x15-3p-5f} & $\bm{0.20 \pm 0.02}$ & $0.12 \pm 0.02$ & $0.03 \pm 0.00$ & $0.08 \pm 0.01$ \\
      && \textbf{15x15-4p-3f} & $\bm{0.41 \pm 0.06}$ & $0.25 \pm 0.07$ & $0.04 \pm 0.01$ & $0.19 \pm 0.02$ \\
      && \textbf{15x15-4p-5f} & $\bm{0.30 \pm 0.02}$ & $0.23 \pm 0.04$ & $0.04 \pm 0.00$ & $0.15 \pm 0.02$ \\
      \midrule
      \multirow{4}{*}{\rotatebox[origin=c]{90}{SMAC}}
      &\multirow{2}{*}{\rotatebox[origin=c]{90}{ret}} & 2s-vs-1sc & $17.83 \pm 0.16${*} & $17.88 \pm 0.74${*} & $17.78 \pm 1.26${*} & $\bm{18.21 \pm 0.39}$ \\
      && 3s-vs-5z & $\bm{16.03 \pm 1.58}$ & $9.66 \pm 2.62$ & $14.11 \pm 2.36${*} & $11.74 \pm 1.87$ \\
      \cmidrule{2-7}
      &\multirow{2}{*}{\rotatebox[origin=c]{90}{win}} & 2s-vs-1sc & $0.79 \pm 0.01$ & $\bm{0.83 \pm 0.04}$ & $0.82 \pm 0.08${*} & $\bm{0.83 \pm 0.02}$ \\
      && 3s-vs-5z & $\bm{0.45 \pm 0.09}$ & $0.08 \pm 0.14$ & $0.29 \pm 0.12${*} & $0.13 \pm 0.11$ \\
      \bottomrule
    \end{tabular}
\end{table*}

\subsection{Evaluation Setup}\label{sec:exp-eval-setup}
We perform evaluation on nine tasks over three benchmark environments.
The tasks can be categorized into two sets according to their challenges:
(1) Challenging sparse-reward tasks focused on efficient exploration. This includes \textsl{SparseTag} and \textsl{Sparse Spread} from MPE, and four tasks with different configurations from LBF.
(2) Tasks that generally assess multi-agent coordination. This includes \textsl{Adversary} in MPE, and an easy task \textsl{2s-vs-1sc} and a hard task \textsl{3s-vs-5z} in SMAC.
Note that LBF tasks and \textsl{Adversary} are fully observable, whereas SMAC and other MPE domains are partially observable environments.
More detailed descriptions of the environments and the evaluation protocol can be found in
\cref{apx:exp-environ} and \cref{apx:eval-protocol}, respectively.

It is important to note that COE is applicable to any value decomposition approach.
To promote fair comparisons in our experiments, we build all exploration methods on the QMIX backbone \citep{rashid2018qmix} with the same network architecture and configurations.
For the same reason, we implement the canonical version of all methods where the only different component is the exploration module unless otherwise specified.
We follow the same protocol presented by \citet{papoudakis2020benchmarking} to optimize hyperparameters.
Specifically we sweep hyperparameters on one task of each environment with three random seeds, and run the best configuration for all tasks in the respective environment with five seeds for the final experiments.
\cref{apx:hyperparam} explains the hyperparameter optimization in more detail.

\begin{figure}[!h]
  \centering
  \includegraphics[width=0.95\linewidth]{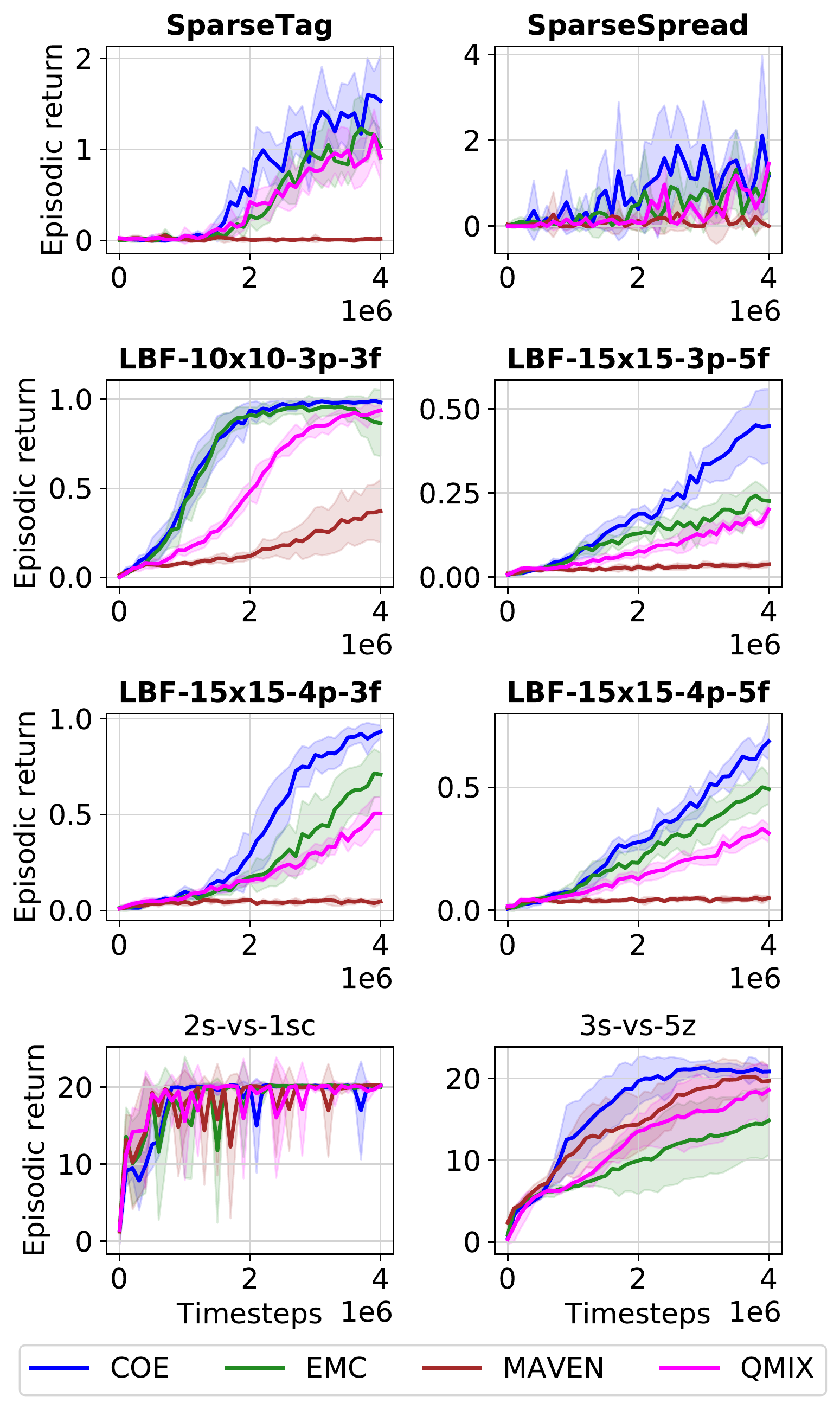}
  \caption{Episodic Returns and 95\% Confidence Interval for All Algorithms in All Tasks except \textsl{Adversary}, with Sparse-Reward Tasks Marked in Bold.}\label{fig:marl-learning-curves-baseline}
\end{figure}

\subsection{Performance}\label{sec:exp-result}
We evaluate COE and compare it with the following state-of-the-art baselines in the experiments:
(\rom{1}) QMIX \citep{rashid2018qmix}: $\varepsilon$-greedy QMIX with linearly annealed epsilon schedule;
(\rom{2}) EMC \citep{zheng2021episodic};
(\rom{3}) MAVEN \citep{mahajan2019maven}: combined with annealing $\varepsilon$-greedy.
Empirical results show that COE outperforms all baselines in the sparse-rewards tasks well-known for exploration challenges, and matches strong baseline performance in general multi-agent tasks.

\cref{tab:avg-return-baseline} summarizes the average returns for the four algorithms in all nine tasks.
The maximum average return is highlighted in bold.
We perform a two-sample t-test \citep{snedecor1980statistical} with a significance level $0.05$ between the best performing algorithm and each of the other algorithms in every task.
The return values are marked with an asterisk if the corresponding algorithm achieves a performance level that is not statistically significantly different from the highest performance.
Difficult exploration tasks are shown in bold.
The same table also reports the average win-rates in SMAC tasks as it is a common practice in MARL literature.
In addition to average returns, the table summarizing maximum returns over training is presented in 
\cref{apx:more-results}.

The results in \cref{tab:avg-return-baseline} and \cref{fig:marl-learning-curves-baseline} show that COE significantly outperforms other baselines in sparse-reward tasks that require efficient exploration.
Particularly COE has higher sample efficiency in difficult LBF domains.
In the very early exploration stage, all algorithms gain performance slowly, resulting in indistinguishable learning curves.
As time progresses, COE makes improvements much faster than the baselines.
The sample efficiency improvement leads to higher final and overall return values.
In relatively easier exploration tasks \textsl{SparseTag}, \textsl{SparseSpread}, and \textsl{Foraging-10x10-3p-3f}, COE's outperformance is not as large as it is in the hard tasks.
Some other baselines also learn strong policies in these tasks.
Since all algorithms are built on the same QMIX agent, overall the results in sparse-reward domains demonstrate the effectiveness of conditional-optimism guided exploration.

In the general MARL coordination tasks, COE has similar performance to the baselines.
\textsl{Adversary} is evidently the easiest task among all tested tasks, where all algorithms quickly converge to the optimal policy at almost identical speed.
In the hard \textsl{3s-vs-5z} task in SMAC, COE shows better sample efficiency and final performance in terms of the mean episodic returns.
This trend is similar to the trends we observe in sparse-reward tasks, although in this task the outperformance is not statistically significant.
These results indicate that COE is not only an effective approach to hard exploration tasks; it is also a strong algorithm generally applicable to common MARL domains.

\begin{table*}[!htb]
    \scriptsize
    \centering
    \caption{Average Returns and 95\% Confidence Interval for All Ablations, and Average Win-rates for SMAC Tasks.}\label{tab:avg-return-ablation}
    \begin{tabular}{*{3}{l}*{5}{c}}
      \toprule
      && \multicolumn{1}{l}{\textbf{Tasks \textbackslash Algs.}} & \multicolumn{1}{c}{COE} & \multicolumn{1}{c}{COE-Cond-IQ} & \multicolumn{1}{c}{COE-Cond-CQ} & \multicolumn{1}{c}{UCB-Ind} & \multicolumn{1}{c}{UCB-Cen} \\
      \midrule
      \multirow{3}{*}{\rotatebox[origin=c]{90}{MPE}}
      && Adversary & $17.77 \pm 0.71$ & $15.46 \pm 0.68$ & $18.99 \pm 0.26$ & $17.70 \pm 0.37$ & $17.15 \pm 0.82$ \\
      && \textbf{Sparse Tag} & $0.65 \pm 0.09$ & $0.07 \pm 0.01$ & $0.83 \pm 0.17$ & $0.52 \pm 0.13$ & $0.49 \pm 0.13$ \\
      && \textbf{Sparse Spread} & $0.79 \pm 0.09$ & $0.36 \pm 0.05$ & $0.54 \pm 0.19$ & $0.59 \pm 0.31$ & $0.75 \pm 0.16$ \\
      \midrule
      \multirow{4}{*}{\rotatebox[origin=c]{90}{LBF}}
      && \textbf{10x10-3p-3f} & $0.71 \pm 0.05$ & $0.76 \pm 0.04$ & $0.68 \pm 0.01$ & $0.67 \pm 0.07$ & $0.64 \pm 0.02$ \\
      && \textbf{15x15-3p-5f} & $0.20 \pm 0.02$ & $0.19 \pm 0.02$ & $0.12 \pm 0.05$ & $0.15 \pm 0.03$ & $0.13 \pm 0.05$ \\
      && \textbf{15x15-4p-3f} & $0.41 \pm 0.06$ & $0.47 \pm 0.06$ & $0.24 \pm 0.06$ & $0.23 \pm 0.05$ & $0.16 \pm 0.10$ \\
      && \textbf{15x15-4p-5f} & $0.30 \pm 0.02$ & $0.23 \pm 0.04$ & $0.14 \pm 0.02$ & $0.23 \pm 0.07$ & $0.27 \pm 0.06$ \\
      \midrule
      \multirow{4}{*}{\rotatebox[origin=c]{90}{SMAC}}
      &\multirow{2}{*}{\rotatebox[origin=c]{90}{ret}} & 2s-vs-1sc & $17.83 \pm 0.16$ & $16.67 \pm 1.56$ & $18.64 \pm 0.48$ & $11.09 \pm 7.27$ & $15.77 \pm 1.45$ \\
      && 3s-vs-5z & $16.03 \pm 1.58$ & $16.85 \pm 1.55$ & $17.01 \pm 0.74$ & $11.03 \pm 3.03$ & $13.36 \pm 3.41$ \\
      \cmidrule{2-8}
      &\multirow{2}{*}{\rotatebox[origin=c]{90}{win}} & 2s-vs-1sc & $0.79 \pm 0.01$ & $0.66 \pm 0.13$ & $0.87 \pm 0.04$ & $0.47 \pm 0.35$ & $0.66 \pm 0.07$ \\
      && 3s-vs-5z & $0.45 \pm 0.09$ & $0.56 \pm 0.15$ & $0.57 \pm 0.06$ & $0.19 \pm 0.18$ & $0.25 \pm 0.21$ \\
      \bottomrule
    \end{tabular}
\end{table*}

\subsection{Ablations}\label{sec:exp-ablation}
COE consists of two major components, namely the independent Q-value functions learned through centralized training, and the conditional optimism.
In order to have a better understanding of COE, we test several ablation variants to evaluate these two components' contribution to performance gain.
Results suggest that conditional optimism plays a dominant role in performance improvement.
Compared to dependent Q-values conditioned on predecessors' actions, independent Q-values learned through value decomposition also work well with conditional optimism in practice.

To evaluate the contributions of \textit{independent} Q-values in COE, we test the following ablation variants that learn \textit{conditional} Q-values: \\
(1) \textbf{COE-Cond-IQ} (conditional optimism + conditional Q, without centralized training): We apply conditional optimism to IQL.
Each agent simultaneously learns an independent Q-network and a dependent Q-network that takes in predecessors' actions as extra inputs.
The dependent network selects actions during training.
Two nets are trained on separate TD losses using the same replay batches.
After training, the independent network is responsible for decision-making at execution time.
This variant directly mimics UCT in MARL without considering each agent's partial observability issue. \\
(2) \textbf{COE-Cond-CQ} (conditional optimism + conditional Q + centralized training): We add a QMIX mixer to COE-Cond-IQ to enable centralized Q-value training. 
The same mixer computes the centralized Q-value $\supidp{\qjt}$ for independent networks and $\supdep{\qjt}$ for dependent networks.

To evaluate the contributions of \textit{conditional} optimism, we propose the following ablation variants that use \textit{non-conditional} optimism: \\
(1) \textbf{UCB-Ind} (independent optimism + independent Q): Similar to COE, each agent performs UCB-based exploration except that optimism is not conditioned on other agents' actions. \\
(2) \textbf{UCB-Cen} (centralized optimism + independent Q): Agents receive UCB optimism only through the intrinsic reward $\frac{c_\subrew}{\sqrt{N(s, \jointaction)}}$ during centralized training.

We follow the same evaluation protocol described in \cref{sec:exp-eval-setup} to conduct experiments.
The average returns of the ablations are summarized in \cref{tab:avg-return-ablation}.
\cref{apx:more-results} and \cref{apx:ablation}
present the learning curves and a more detailed description of the ablations, respectively.
Results show that COE has a similar performance as COE-Cond-CQ and COE-Cond-IQ in the majority of tested tasks.
COE-Cond-IQ performs relatively worse in MPE tasks, but better in LBF tasks.
This may be attributed to the partial observability issue:
since LBF is fully observable, COE-Cond-IQ becomes a more legitimate adoption of UCT to cooperative MARL.
COE-Cond-CQ matches COE's performance in MPE and SMAC.
Although it underperforms COE in three LBF tasks, COE-Cond-CQ is still competitive and matches EMC's performance in LBF.
These results suggest that conditional optimism boosts sample efficiency and overall performance with different Q-value estimation approaches.

On the other hand, UCB-Ind underperforms COE in hard LBF tasks and SMAC tasks.
It also has a large variance across random seeds in SMAC tasks.
UCB-Cen matches COE in half of the tasks, but it also suffers from large variances.
Through these comparisons, we observe conditional optimism guides more steady performance improvement.

\subsection{Didactic Problem}
A rigorous application of tree-based methods in MARL requires learning each agent's state-action value conditioned on earlier agents' state-action pairs.
Due to partial observability in MARL that prevents access to global states, we use value decomposition as an approximate implementation of conditional value estimation and empirically show its effectiveness in previous sections.
In this section, we provide a sanity check on a repeated multi-player game problem to re-demonstrate conditional optimism is important whereas conditional value estimation could be unnecessary.

We consider the cooperative multi-player game problem, where the common payoff is based on the joint action of a group of agents.
Suppose we have $n$ agents, each has $k$ actions.
In our didactic Bernoulli game, only one out of $k^n$ joint actions is optimal with payoff distribution $\mathcal{B}(p=0.9)$, and all other joint actions are sub-optimal with payoff distribution $\mathcal{B}(p=p_0)$, where the sub-optimality value $p_0$ is an environment hyper-parameter.
For each game instance, a uniformly sampled joint action from all combinations is set to be the optimal.

\begin{figure}[!t] 
  \centering
  \includegraphics[width=0.8\linewidth]{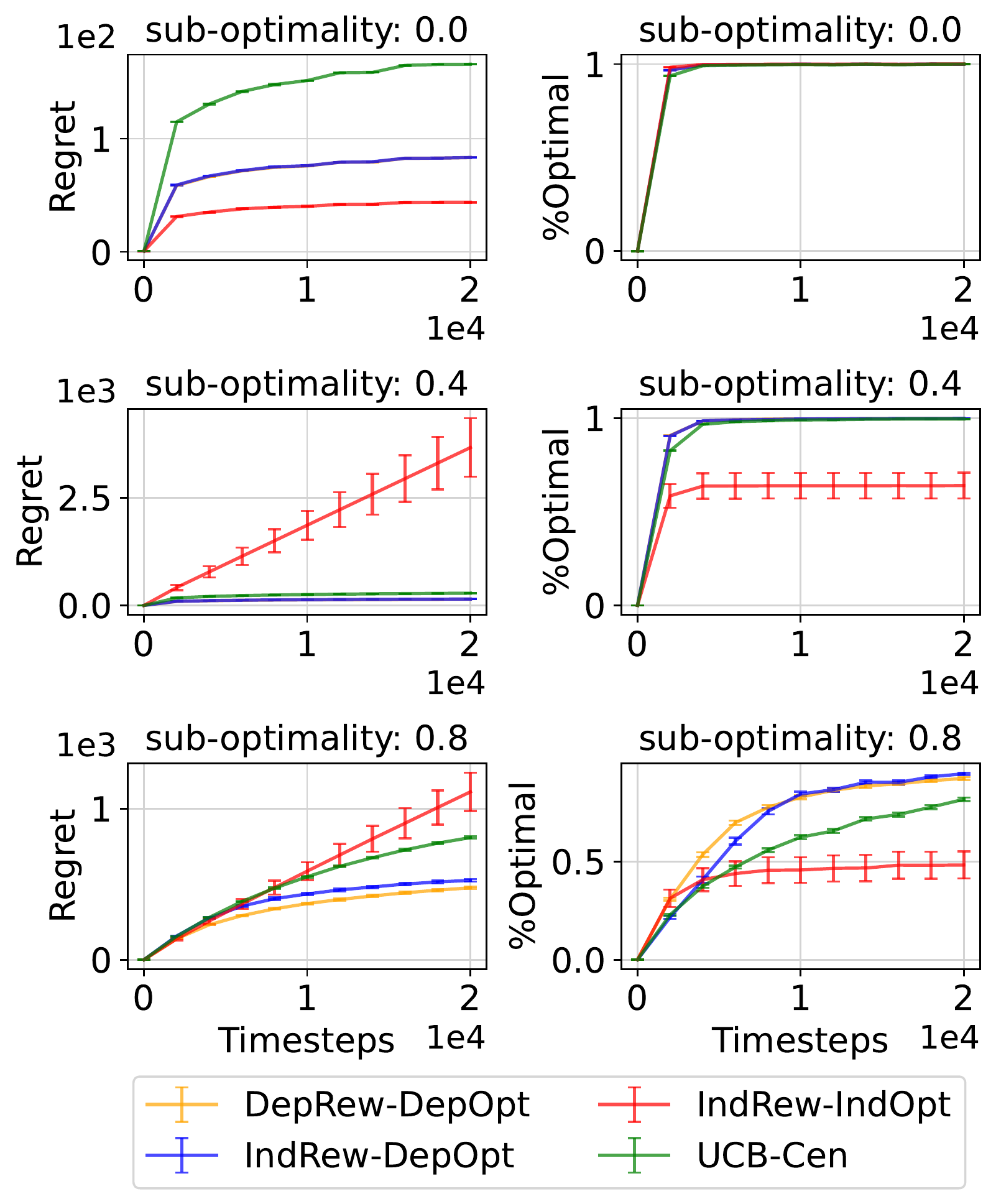}
  \caption{Mean Performance and Standard Error for Multi-player Game Problem with $8$ Agents and $3$ Actions Each.}\label{fig:bandit-learning-curves}
\end{figure}

We test four UCB-based algorithm variants: \\
(1) \textbf{DepRew-DepOpt}: it performs UCT exploration as in \cref{eq:uct-bandit-act}, where both payoff estimates and count-based optimism are dependent on prior agents' joint action; \\
(2) \textbf{IndRew-DepOpt}: each agent maintains its payoff estimates independently, but the optimism is dependent on predecessors; \\
(3) \textbf{IndRew-IndOpt}: both payoff estimates and optimism are independent; \\
(4) \textbf{UCB-Cen}: one UCB learner whose action space is the cartesian product of all agents' action set.
This variant is merely for performance comparison because it is not factorizable to decentralized agents.

We run experiments on a game problem with $8$ agents and $3$ actions of each agent over $50$ seeds, and report the performance of all four algorithms for different sub-optimality settings in \cref{fig:bandit-learning-curves}.
We evaluate algorithms with two metrics, the expected regret shown in the left column, which is preferably lower, and the percentage of optimal joint action being selected shown in the right column, which is preferably higher.
Results show that both DepRew-DepOpt and IndRew-DepOpt quickly converge to the optimal policy across different sub-optimality settings.
Their learning curves overlap when sub-optimality is $0.0$ or $0.4$, and they have only marginal difference when sub-optimality equals $0.8$.
This suggests that in this game task, conditional optimism robustly drives efficient cooperative exploration, regardless of whether the payoff estimates are learned independently or not.
IndRew-IndOpt, on the other hand, is inefficient to identify the optimal joint action and has high variances across random seeds.
These results highlight the significance of conditional count-based optimism, and its dominant role over the action-value estimates in coordinated exploration.
In general, these results are consistent with MARL ablation results from \cref{sec:exp-ablation}.

\section{Conclusions}\label{sec:conclusion}
In this paper, we draw the connection between cooperative multi-agent reinforcement learning (MARL) and tree search.
Inspired by the tree search algorithm UCT, we propose a multi-agent exploration method Conditionally Optimistic Exploration (COE), that utilizes the sequential decision-making scheme and visitation count conditioned on previous agents' actions.
Empirical results show that our method significantly outperforms state-of-the-art MARL baselines in sparse-reward hard-exploration tasks, and matches their performance in general coordination tasks.

One limitation of our method is that it may require a large amount of memory due to storing visitation counts of state-action tuples during training, which makes our method costly to scale to tasks with very large state-action space.
An interesting future work is to utilize neural network density models to estimate pseudo-counts.
Training such a model would require more computation but the model itself only occupies constant memory.


\begin{acknowledgements}
We acknowledge the computational resources provided by the Digital Research Alliance of Canada.
Janarthanan Rajendran acknowledges the support of the IVADO postdoctoral fellowship.
Sarath Chandar acknowledges the support of the Canada CIFAR AI Chair program and an NSERC Discovery Grant.
\end{acknowledgements}


\newpage
\onecolumn
\appendix
\section{Pseudo-count for Deep RL}\label{apx:pseudo-count}
Counting visitations in high-dimensional or continuous state space could be challenging.
This section introduces how we approximate counts by applying the static hashing \citep{tang2017exploration} method, a well-established count approximation approach in RL, adopted successfully in works such as \citet{rashid2020optimistic}.

In particular, the state $s \in \statespace$ is projected to a lower-dimensional feature space by $\phi(s) = sgn (A g(s)) \in \{-1, 1\}^k$, 
where $g: \statespace \rightarrow \mathbb{R}^D$ is an optional pre-processing function,
$A \in \mathbb{R}^{k \times D}$ is a projection matrix with entries drawn i.i.d. from a unit Gaussian distribution $\mathcal{N}(0,1)$,
and $sgn(\cdot)$ is the element-wise sign function.
This method clusters similar states in $\statespace$ to one feature in a small, countable feature space, which enables us to count.
The $k$ value controls the granularity of state approximation: higher $k$ leads to more distinguishable features yet less generalizability across similar states.
We record the visitation count for the tuple of the state feature $\phi(s)$ and all agents' joint action $\jointaction$, denoted by $N(s, \jointaction)$ for simplicity of notation.
Note that for each agent $i$, the count up to its action $a_i$ satisfies:
\begin{align*}
    N(s, a_{<i}, a_i) &= \sum_{a_{i+1}} N(s, a_{<i}, a_i, a_{i+1}) \\
        &= \sum_{a_{>i}} N(s, a_{<i}, a_i, a_{>i}),
\end{align*}
where $a_{<i}$ and $a_{>i}$ denote the joint actions taken by preceding and subsequent agents of $i$, respectively.
This relationship shows that we can obtain any count up to $a_i$ by summing up the counts of joint actions that overlap $a_{<i}$ at state $s$.
This relationship is naturally aligned with the tree structure, where the total count of each node equals the number of action sequences going through that node.
Thus we are able to perform optimistic exploration using conditional counts.

\section{Environment details}\label{apx:exp-environ}

\paragraph{Multi-Agent Particle Environment}
Multi-Agent Particle Environment (MPE) \citep{lowe2017multi,mordatch2018emergence} is a suite of two-dimensional navigation tasks where the entities in the environment obey physics properties.
We choose three tasks that do not involve agent-wide communication: \textsl{Sparse Spread}, \textsl{Sparse Tag}, and \textsl{Adversary}.
In the first two tasks, reward signals are sparse and agents receive positive rewards only when they jointly complete the task.
They are almost fully observable except each agent does not observe the velocity of other agents.
\textsl{Adversary} is fully observable.

\paragraph{Level-Based Foraging}
Level-Based Foraging (LBF) \citep{albrecht2015game,christianos2020shared,papoudakis2020benchmarking} is a set of food-collection tasks in a grid-world.
Each agent or food item is assigned a level value, such that a group of agents can pick up a food item if the sum of agents' levels is greater than or equal to the item's level.
Agents receive a positive reward only when a food item is picked up, hence LBF requires efficient coordinated exploration.
We choose four tasks with different grid dimensions, number of agents, and number of food items.
By default, they are all fully observable.

\paragraph{StarCraft Multi-Agent Challenge}
StarCraft Multi-Agent Challenge (SMAC) \citep{samvelyan2019starcraft} consists of battle tasks where a group of agents is learned to defeat another group.
Each agent could only observe entities within a fixed-sized window.
All tasks have dense rewards, and agents start engaging immediately after the game starts.
As \citet{mahajan2019maven} point out, SMAC tasks are not designed to evaluate cooperative exploration.
In order to assess coordination in partially-observable and non-stationary settings, we choose one easy task \textit{2s-vs-1sc} and one hard task \textit{3s-vs-5z}.

\section{Evaluation Protocol}\label{apx:eval-protocol}
In each task we train all algorithms for four million timesteps.
During training we perform $41$ evaluations at constant timestep intervals, that is, 100k timestep intervals, and at each evaluation point we evaluate for 100 episodes.
We train each algorithm with parameter sharing, where all agent networks share the same set of parameters, and the one-hot identity of each agent as additional network input helps the neural network to develop diverse behaviour.

We evaluate algorithms' performance in a task by two metrics: maximum returns and average returns.
The maximum return refers to the highest mean evaluation return across five seeds achieved at one evaluation point during training.
This metric evaluates algorithms' best-reached performance in a task
The average return is the evaluation return averaged over all evaluation points during training.
This metric reflects both sample efficiency and final performance.

\section{Additional results}\label{apx:more-results}

\begin{figure*}[!htb]
  \centering
  \includegraphics[width=0.9\linewidth]{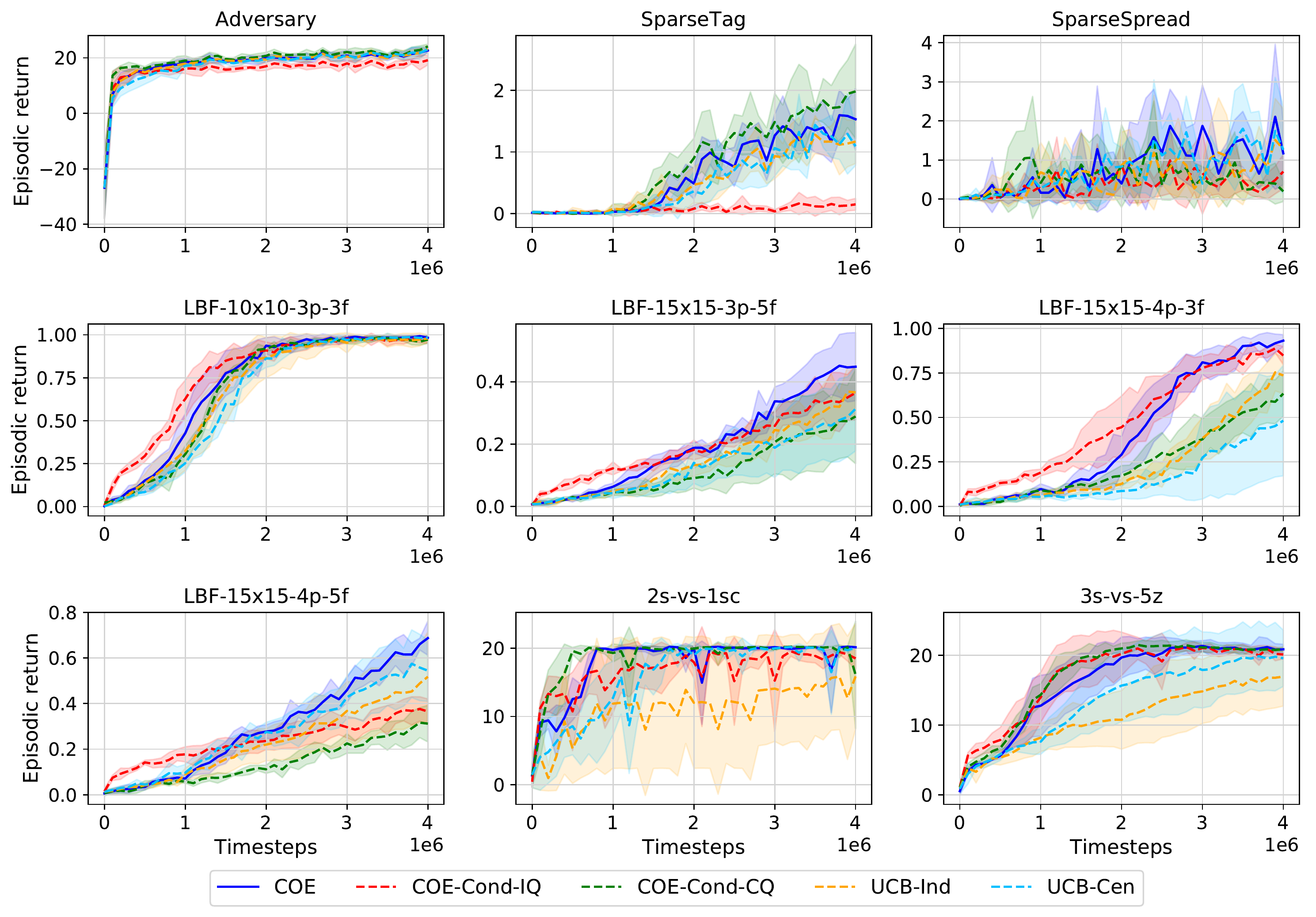}
  \caption{Episodic Returns and 95\% Confidence Interval for All Ablations in All Tasks.}\label{fig:marl-learning-curves-ablation}
\end{figure*}

Table~\ref{tab:max-return-all} summarizes the \textit{maximum} returns for all eight algorithms (including the ablations) in all nine tasks,
which also reports the maximum win-rates in SMAC tasks.
Figure~\ref{fig:marl-learning-curves-ablation} presents learning curves of the evaluation returns achieved during training by ablations in all nine tasks.
Sparse-reward tasks have bold titles.

\begin{table*}[!htb]
    \tiny
    \centering
    \caption{Maximum Returns and 95\% Confidence Interval for All Eight Algorithms in All Nine Tasks, and Maximum Win-rates for SMAC Tasks.}\label{tab:max-return-all}
    \begin{tabular}{*{3}{l}*{8}{c}}
      \toprule 
      && \multicolumn{1}{l}{\textbf{Tasks \textbackslash Algs.}} & \multicolumn{1}{c}{COE} & \multicolumn{1}{c}{COE-Cond-IQ} & \multicolumn{1}{c}{COE-Cond-CQ} & \multicolumn{1}{c}{UCB-Ind} & \multicolumn{1}{c}{UCB-Cen} & \multicolumn{1}{c}{EMC} & \multicolumn{1}{c}{MAVEN} & \multicolumn{1}{c}{QMIX} \\
      \midrule
      \multirow{3}{*}{\rotatebox[origin=c]{90}{MPE}}
      && Adversary & $22.68 \pm 0.80$ & $19.18 \pm 1.70$ & $24.14 \pm 0.83$ & $23.16 \pm 1.28$ & $23.02 \pm 0.93$  & $22.03 \pm 2.12$  & $23.52 \pm 1.50$  & $22.70 \pm 1.61$  \\
      && Sparse Tag & $1.60 \pm 0.41$ & $0.16 \pm 0.18$ & $1.98 \pm 0.77$ & $1.28 \pm 0.31$ & $1.44 \pm 0.05$  & $1.23 \pm 0.35$  & $0.06 \pm 0.03$ & $1.16 \pm 0.29$  \\
      && Sparse Spread & $2.11 \pm 1.86$ & $0.99 \pm 0.85$ & $1.46 \pm 1.05$ & $1.51 \pm 1.06$ & $1.80 \pm 1.15$  & $1.31 \pm 0.92$  & $0.43 \pm 0.85$  & $1.46 \pm 0.28$  \\
      \midrule
      \multirow{4}{*}{\rotatebox[origin=c]{90}{LBF}}
      && 10x10-3p-3f & $0.99 \pm 0.01$ & $0.98 \pm 0.02$ & $0.98 \pm 0.01$ & $0.98 \pm 0.02$ & $0.99 \pm 0.01$ & $0.96 \pm 0.04$  & $0.37 \pm 0.18$ & $0.94 \pm 0.03$ \\
      && 15x15-3p-5f & $0.45 \pm 0.10$ & $0.36 \pm 0.09$ & $0.29 \pm 0.15$ & $0.37 \pm 0.08$ & $0.31 \pm 0.14$  & $0.24 \pm 0.04$ & $0.04 \pm 0.01$ & $0.20 \pm 0.02$ \\
      && 15x15-4p-3f & $0.93 \pm 0.03$ & $0.89 \pm 0.02$ & $0.63 \pm 0.13$ & $0.75 \pm 0.11$ & $0.48 \pm 0.31$ & $0.71 \pm 0.13$ & $0.06 \pm 0.01$ & $0.51 \pm 0.09$ \\
      && 15x15-4p-5f & $0.69 \pm 0.08$ & $0.38 \pm 0.05$ & $0.32 \pm 0.07$ & $0.52 \pm 0.20$ & $0.57 \pm 0.15$  & $0.50 \pm 0.08$ & $0.05 \pm 0.01$ & $0.33 \pm 0.04$ \\
      \midrule
      \multirow{4}{*}{\rotatebox[origin=c]{90}{SMAC}}
      &\multirow{2}{*}{\rotatebox[origin=c]{90}{ret}} & 2s-vs-1sc & $20.25 \pm 0.01$ & $19.57 \pm 0.73$ & $20.24 \pm 0.00$ & $15.88 \pm 7.79$ & $20.19 \pm 0.07$  & $20.22 \pm 0.06$ & $20.22 \pm 0.04$  & $20.16 \pm 0.05$ \\
      && 3s-vs-5z & $21.32 \pm 0.75$ & $21.16 \pm 0.56$ & $21.47 \pm 0.59$ & $16.93 \pm 4.24$ & $19.86 \pm 5.03$  & $14.84 \pm 4.19$ & $20.15 \pm 1.43$  & $18.57 \pm 3.01$  \\
      \cmidrule{2-11}
      &\multirow{2}{*}{\rotatebox[origin=c]{90}{win}} & 2s-vs-1sc & $1.00 \pm 0.00$ & $0.92 \pm 0.09$ & $1.00 \pm 0.00$ & $0.77 \pm 0.38$ & $0.99 \pm 0.01$  & $1.00 \pm 0.00$ & $1.00 \pm 0.00$ & $0.99 \pm 0.00$ \\
      && 3s-vs-5z & $0.97 \pm 0.00$ & $0.93 \pm 0.05$ & $0.98 \pm 0.02$ & $0.56 \pm 0.45$ & $0.61 \pm 0.37$  & $0.27 \pm 0.37$ & $0.87 \pm 0.16$  & $0.65 \pm 0.30$  \\
      \bottomrule
    \end{tabular}
\end{table*}

\section{Ablation details}\label{apx:ablation}

In this section, we present in detail the ablation variants
introduced in \cref{sec:exp-ablation}.

COE-Cond-IQ directly adopts the idea of UCT, without considering the partial observability issue of each agent.
In order to enable decentralized execution, we simultaneously learn a Q-value function dependent on preceding agents' actions and its independent counterpart.
Similar to the MACPF factorization \citep{wang2022more}, each agent $i$ has an independent Q-network $\supidp{\qi} (\tau_i, a_i; \phi_i)$ parameterized by $\phi_i$,
and a dependency correction network $\supdep{c}_i(\tau_i, a_i | a_{<i}; \psi_i)$ parameterized by $\psi_i$,
whose sum constructs the dependent Q-network $\supdep{\qi}(\tau_i, a_i | a_{<i}; \phi_i, \psi_i) = \supidp{\qi} (\tau_i, a_i; \phi_i) + \supdep{c}_i(\tau_i, a_i | a_{<i}; \psi_i)$.

\begin{figure*}[!tb] 
  \centering
  \includegraphics[width=1.0\linewidth]{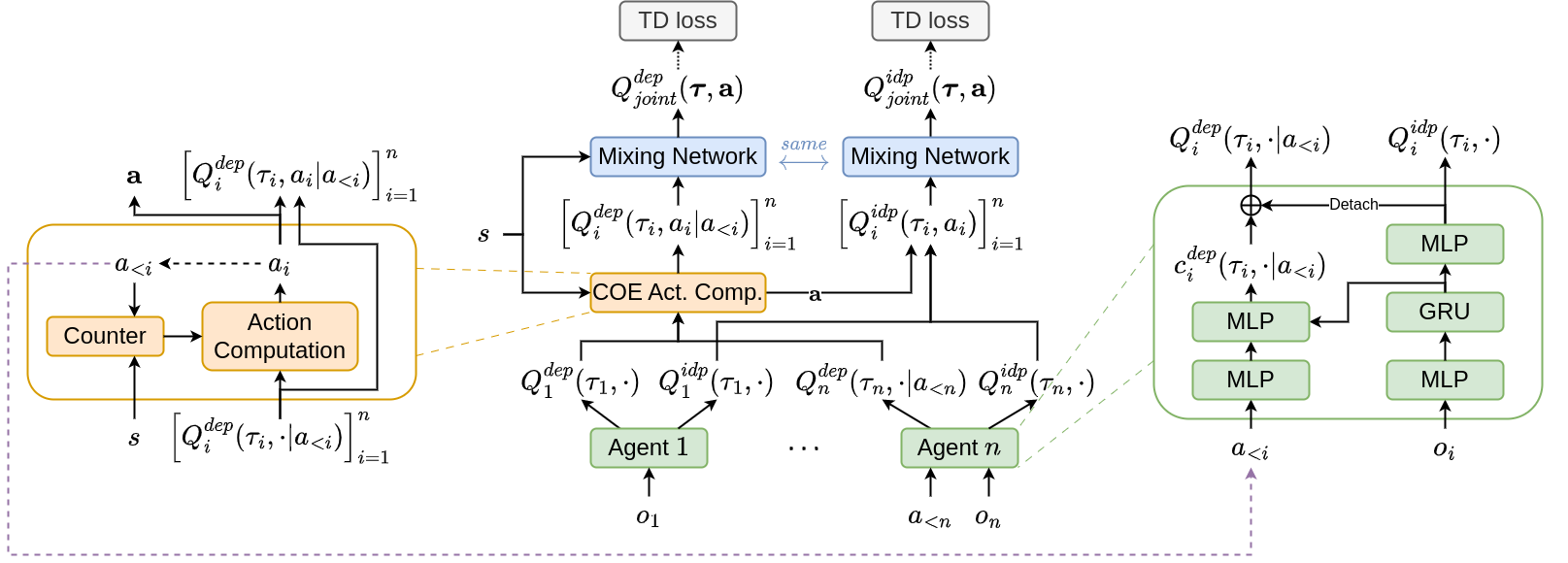}
  \caption{Learning Framework for COE-Cond-CQ.}\label{fig:coe-cond-cq-framework}
\end{figure*}

Individual agent's action-value networks $\supdep{\qi}$ and $\supidp{\qi}$ are separately trained by minimizing the mean-squared TD error on each Q-network:
\begin{align}
    \supdep{\mathcal{L}}_i( \psi_i ) &= \EE_\replay [ (\supdep{\qi}(\tau_i, a_i | a_{<i}) - \supdep{y}_i )^2 ] \label{eq:td-err-dep-i} \\
    \supidp{\mathcal{L}}_i( \phi_i ) &= \EE_\replay [ (\supidp{\qi}(\tau_i, a_i) - \supidp{y}_i )^2 ] \label{eq:td-err-ind-i}
\end{align}
where $\supdep{y}_i = (r + \gamma \max_{a_i'} (\supdep{\qi}(\tau_i', a_i' | a_{<i}) ) )$ and $\supidp{y}_i = (r + \gamma \max_{a_i'} (\supidp{\qi}(\tau_i', a_i') ) )$ are the update targets, and $\replay$ contains trajectory data collected by $\supdep{\qi}$'s.
To ensure $\supdep{\qi}$ and $\supidp{\qi}$ achieve the same performance, they are constructed and trained in a way that strengthens their coupling:
$\supdep{\qi}$ is the combination of $\supidp{\qi}$ and a correction network;
during training the same mini-batch of trajectory data sampled from $\replay$ is used to compute both $\supdep{\mathcal{L}}_i$ and $\supidp{\mathcal{L}}_i$.

COE exploration is applied to this variant in a similar way as being applied to value decomposition methods.
The optimistic bonus is added to $\supdep{\qi}$ at action selection during training.
Note that for each agent $i$ the optimistic TD update target is applied to both \cref{eq:td-err-dep-i} and \cref{eq:td-err-ind-i}:
\begin{align}\label{eq:ucb-target-i}
    & y_i = \left( r(s, \jointaction) + \frac{c_\subrew}{\sqrt{N(s, a_{<i}, a_i)}} \right)
    + \gamma \max_{a_i'} \left( Q_i(\tau_i', a_i') + \frac{c_\subboot}{\sqrt{N(s', a_{<i}', a_i')}} \right),
\end{align}
where $c_\subrew, c_\subboot \in \mathbb{R}_+$ are hyper-parameters controlling the scale of the optimistic bias in reward and bootstrapped target, respectively.
During decentralized execution, agents take actions according to $\supidp{\qi}$'s only.

We name this variant COE-Cond-IQ as it could be considered as a direct adoption of UCT to IQL \citep{tan1993multi}.
As opposed to the utility function that learns implicit dependency via centralized training in value decomposition methods, each agent learns a Q-value function, that explicitly captures the correlation among agents by conditioning on previous agents' actions.
COE-Cond-IQ also complies with the CTDE paradigm.
However, it ignores the partial observability of each individual agent.
Each agent only has access to its own local trajectory history.

Another ablation we introduce is COE-Cond-CQ, which combines centralized training and COE-Cond-IQ.
The learning framework of COE-Cond-CQ is illustrated in \cref{fig:coe-cond-cq-framework}.
The same mixing network $\mixer(\cdot ; \theta)$ we use in COE is used to compute both dependent and independent joint Q-values:
\begin{align}
    & \supdep{\qjt}(\jointhist, \jointaction) = \mixer \left( [ \supdep{\qi}(\tau_i, a_i | a_{<i}) ]^N_{i=1}, s; \theta \right) \label{eq:q-joint-dep-condcq} \\
    & \supidp{\qjt}(\jointhist, \jointaction) = \mixer \left( [ \supidp{\qi}(\tau_i, a_i) ]^N_{i=1}, s; \theta \right) \label{eq:q-joint-ind-condcq}
\end{align}
Similarly, centralized training also optimizes both dependent and independent mean-squared TD error:
\begin{align}
    \supdep{\mathcal{L}}( [\psi]^N_{i=1}, \theta ) = \EE_\replay [ (\supdep{\qjt}(\jointhist, \jointaction) - \supdep{y} )^2 ] \label{eq:td-err-dep-condcq} \\
    \supidp{\mathcal{L}}( [\phi]^N_{i=1}, \theta ) = \EE_\replay [ (\supidp{\qjt}(\jointhist, \jointaction) - \supidp{y} )^2 ] \label{eq:td-err-ind-condcq}
\end{align}
where $\supdep{y} = (r + \gamma \max_{\jointaction'} (\supdep{\qjt}(\jointhist', \jointaction') ) )$ and $\supidp{y} = (r + \gamma \max_{\jointaction'} (\supidp{\qjt}(\jointhist', \jointaction') ) )$ are update targets for dependent and independent networks, respectively.
Exploration is performed the same way as COE, and action selection is performed the same way as COE-Cond-IQ.

In the ablation UCB-Ind, each agent performs UCB-based exploration independently.
It is straightforward to obtain UCB-Ind: we simply replace any conditional count terms in COE with independent counts, which do not rely on other agents' actions.

The ablation UCB-Cen augments the global reward with an intrinsic reward $\frac{c_\subrew}{\sqrt{N(s, \jointaction)}}$.
Agents learn optimistic Q-values through centralized training.

\section{Hyperparameter settings}\label{apx:hyperparam}

To perform hyperparameter optimization we follow the same protocol presented by \citet{papoudakis2020benchmarking}.
We select one task from each benchmark environment and optimize the hyperparameters of all algorithms in this task.
In particular, we select \textit{Sparse Tag} from MPE, \textit{15x15-3p-5f} from LBF, and \textit{3s-vs-5z} from SMAC.
We perform a coarse grid search on hyperparameter settings and train each configuration with three seeds.
We identify the best configuration according to the maximum evaluation returns.
This best configuration on each task is then used for all tasks in the respective environment for the final experiments with five seeds.

For methods that use intrinsic reward --- i.e. COE, EMC, and MAVEN --- we only test constant intrinsic reward scales.
For COE, the hyperparameter combination with $c_\subact = c_\subrew = c_\subboot = 0$ is ignored as this setting refers to the greedy-action QMIX.
For MAVEN, we determine the hyperparameter settings according to the original paper and its accompanying codebase.
In particular, we sweep the intrinsic scales only when "MI intrinsic" is True.
The hyperparameters "MI intrinsic" and "RNN discriminator" cannot both be True.
When MAVEN uses $\varepsilon$-greedy, the epsilon annealing time is 50k timesteps.
Every epsilon annealing schedule --- utilized by either MAVEN or QMIX --- is linear with an initial value of $1.0$ and a final value of $0.0$.

\begin{table}[!htb] 
    \centering
    \caption{Common QMIX Hyperparameters for All algorithms across All Tasks.}\label{tab:apx-hyperparam-common}
    \begin{tabular}{cc}
      \toprule 
      Hyperparameter Name & Value \\
      \midrule
      hidden dimension & 128 \\
      reward standardization & True \\
      network type & GRU \\
      evaluation epsilon & 0 \\
      target update & 0.01 (soft) \\
      \bottomrule
    \end{tabular}
\end{table}

\begin{table}[!htb]
    \centering
    \caption{Hyperparameters for COE: Values Swept in Grid-search and Best Configuration for each Benchmark.}\label{tab:apx-hyperparam-coe}
    \begin{tabular}{*{5}{c}}
      \toprule 
      Hyperparameter Name & Swept values & MPE & LBF & SMAC \\
      \midrule
      learning rate & 0.0001/0.0003/0.0005 & 0.0001 & 0.0003 & 0.0005 \\
      feature dimension $k$ & 8/12/16 & 8 & 16 & 8 \\
      $c_\subact$ & 0/0.01/0.05 & 0.01 & 0.01 & 0 \\
      $c_\subboot$ & 0/0.01/0.05 & 0 & 0 & 0 \\
      $c_\subrew$ & 0/0.01/0.05 & 0.05 & 0 & 0.05 \\
      \bottomrule
    \end{tabular}
\end{table}

\begin{table}[!htb]
    \centering
    \caption{Hyperparameters for COE-Cond-IQ: Values Swept in Grid-search and Best Configuration for each Benchmark.}\label{tab:apx-hyperparam-coe-cond-iq}
    \begin{tabular}{*{5}{c}}
      \toprule 
      Hyperparameter Name & Swept values & MPE & LBF & SMAC \\
      \midrule
      learning rate & 0.0001/0.0003/0.0005 & 0.0001 & 0.0001 & 0.0005 \\
      feature dimension $k$ & 8/12/16 & 8 & 8 & 16 \\
      $c_\subact$ & 0/0.01/0.05 & 0 & 0.05 & 0.01 \\
      $c_\subboot$ & 0/0.01/0.05 & 0 & 0 & 0 \\
      $c_\subrew$ & 0/0.01/0.05 & 0.05 & 0 & 0 \\
      \bottomrule
    \end{tabular}
\end{table}

\begin{table}[!htb]
    \centering
    \caption{Hyperparameters for COE-Cond-CQ: Values Swept in Grid-search and Best Configuration for each Benchmark.}\label{tab:apx-hyperparam-coe-cond-cq}
    \begin{tabular}{*{5}{c}}
      \toprule 
      Hyperparameter Name & Swept values & MPE & LBF & SMAC \\
      \midrule
      learning rate & 0.0001/0.0003/0.0005 & 0.0001 & 0.0003 & 0.0005 \\
      feature dimension $k$ & 8/12/16 & 12 & 16 & 8 \\
      $c_\subact$ & 0/0.01/0.05 & 0.05 & 0.01 & 0.05 \\
      $c_\subboot$ & 0/0.01/0.05 & 0 & 0 & 0.01 \\
      $c_\subrew$ & 0/0.01/0.05 & 0.05 & 0.01 & 0.01 \\
      \bottomrule
    \end{tabular}
\end{table}

\begin{table}[!htb]
    \centering
    \caption{Hyperparameters for UCB-Indep: Values Swept in Grid-search and Best Configuration for each Benchmark.}\label{tab:apx-hyperparam-ucb-indep}
    \begin{tabular}{*{5}{c}}
      \toprule 
      Hyperparameter Name & Swept values & MPE & LBF & SMAC \\
      \midrule
      learning rate & 0.0001/0.0003/0.0005 & 0.0001 & 0.0003 & 0.0005 \\
      feature dimension $k$ & 8/12/16 & 8 & 12 & 12 \\
      $c_\subact$ & 0/0.01/0.05 & 0.01 & 0.01 & 0 \\
      $c_\subboot$ & 0/0.01/0.05 & 0 & 0.01 & 0 \\
      $c_\subrew$ & 0/0.01/0.05 & 0 & 0.01 & 0.01 \\
      \bottomrule
    \end{tabular}
\end{table}

\begin{table}[!htb]
    \centering
    \caption{Hyperparameters for UCB-Central: Values Swept in Grid-search and Best Configuration for each Benchmark.}\label{tab:apx-hyperparam-ucb-central}
    \begin{tabular}{*{5}{c}}
      \toprule 
      Hyperparameter Name & Swept values & MPE & LBF & SMAC \\
      \midrule
      learning rate & 0.0001/0.0003/0.0005 & 0.0001 & 0.0003 & 0.0005 \\
      feature dimension $k$ & 8/12/16 & 8 & 8 & 16 \\
      $c_\subrew$ & 0/0.01/0.05 & 0.05 & 0.01 & 0.05 \\
      \bottomrule
    \end{tabular}
\end{table}

\begin{table}[!htb]
    \centering
    \caption{Hyperparameters for EMC: Values Swept in Grid-search and Best Configuration for each Benchmark.}\label{tab:apx-hyperparam-emc}
    \begin{tabular}{*{5}{c}}
      \toprule 
      Hyperparameter Name & Swept values & MPE & LBF & SMAC \\
      \midrule
      learning rate & 0.0001/0.0003/0.0005 & 0.0001 & 0.0003 & 0.0005 \\
      curiosity scale & 0.001/0.005/0.01/0.05/0.1/0.5/1.0 & 0.01 & 0.001 & 0.001 \\
      \bottomrule
    \end{tabular}
\end{table}

\begin{table}[!htb]
    \centering
    \caption{Hyperparameters for MAVEN: Values Swept in Grid-search and Best Configuration for each Benchmark.}\label{tab:apx-hyperparam-maven}
    \begin{tabular}{*{5}{c}}
      \toprule 
      Hyperparameter Name & Swept values & MPE & LBF & SMAC \\
      \midrule
      learning rate & 0.0001/0.0003/0.0005 & 0.0003 & 0.0001 & 0.0005 \\
      RNN discriminator & True/False & False & False & False \\
      MI intrinsic & True/False & True & True & True \\
      curiosity scale & 0.001/0.005/0.01/0.05/0.1/0.5/1.0 & 0.01 & 0.05 & 0.005 \\
      noise bandit & True/False & True & False & True \\
      epsilon start & 0.0/1.0 & 1.0 & 1.0 & 1.0 \\
      \bottomrule
    \end{tabular}
\end{table}

\begin{table}[!htb]
    \centering
    \caption{Hyperparameters for QMIX: Values Swept in Grid-search and Best Configuration for each Benchmark.}\label{tab:apx-hyperparam-qmix}
    \begin{tabular}{*{5}{c}}
      \toprule 
      Hyperparameter Name & Swept values & MPE & LBF & SMAC \\
      \midrule
      learning rate & 0.0001/0.0003/0.0005 & 0.0001 & 0.0001 & 0.0005 \\
      epsilon anneal & 50,000/200,000 & 50,000 & 50,000 & 50,000 \\
      \bottomrule
    \end{tabular}
\end{table}

\end{document}